\documentclass[10pt,twocolumn,letterpaper]{article}

\usepackage[pagenumbers]{cvpr} 

\usepackage{graphicx}
\usepackage{amsmath}
\usepackage{amssymb}
\usepackage{booktabs}
\usepackage{tabularx}
\usepackage{multirow}
\newcolumntype{L}[1]{>{\raggedright\arraybackslash}p{#1}}
\newcolumntype{C}[1]{>{\centering\arraybackslash}p{#1}}
\newcolumntype{R}[1]{>{\raggedleft\arraybackslash}p{#1}}

\usepackage[pagebackref,breaklinks,colorlinks]{hyperref}

\usepackage[capitalize]{cleveref}
\crefname{section}{Sec.}{Secs.}
\Crefname{section}{Section}{Sections}
\Crefname{table}{Table}{Tables}
\crefname{table}{Tab.}{Tabs.}

\def\ours{\texttt{\textbf{STHG}}}

\def\cvprPaperID{16} 
\def\confName{Ego4D}
\def\confYear{2023}

\begin{document}

\title{\ours: Spatial-Temporal Heterogeneous Graph Learning\\for Advanced Audio-Visual Diarization}

\author{Kyle Min \\\\
Intel Labs\\
{\tt\small kyle.min@intel.com}
}
\maketitle

\begin{abstract}
This report introduces our novel method named \ours{} for the Audio-Visual Diarization task of the Ego4D Challenge 2023. Our key innovation is that we model all the speakers in a video using a single, unified heterogeneous graph learning framework. Unlike previous approaches that require a separate component solely for the camera wearer, \ours{} can jointly detect the speech activities of all people including the camera wearer. Our final method obtains 61.1\% DER on the test set of Ego4D, which significantly outperforms all the baselines as well as last year's winner. Our submission achieved 1st place in the Ego4D Challenge 2023. We additionally demonstrate that applying the off-the-shelf speech recognition system to the diarized speech segments by \ours{} produces a competitive performance on the Speech Transcription task of this challenge.
\end{abstract}

\section{Introduction}
\label{sec:intro}
The main goal of Audio-Visual Diarization (AVD) is to identify ``who speaks when'' in a video. Despite recent advances, state-of-the-art approaches~\cite{cabanas2018multimodal,chung2019said,kang2020multimodal,xu2021ava,sharma2022using} rely on a restrictive assumption that active speakers are always visible in the scene, which does not hold for egocentric videos. In egocentric videos, a camera wearer (CW) is never visible, although its speech activities are frequently recorded. To address this issue, previous approaches use two different methods: one for detecting the speech activities of every visible speaker using their audio-visual input, and the other for detecting the CW's speech activities from its audio information alone~\cite{kang2020multimodal,grauman2022ego4d}. However, this approach fails to jointly model these two types of speakers, as the CW is analyzed independently. Existing AVD methods are hence unable to fully leverage the inter-speaker context between the CW and the others.

\begin{figure}[t!]
\centering
\includegraphics[width=\linewidth]{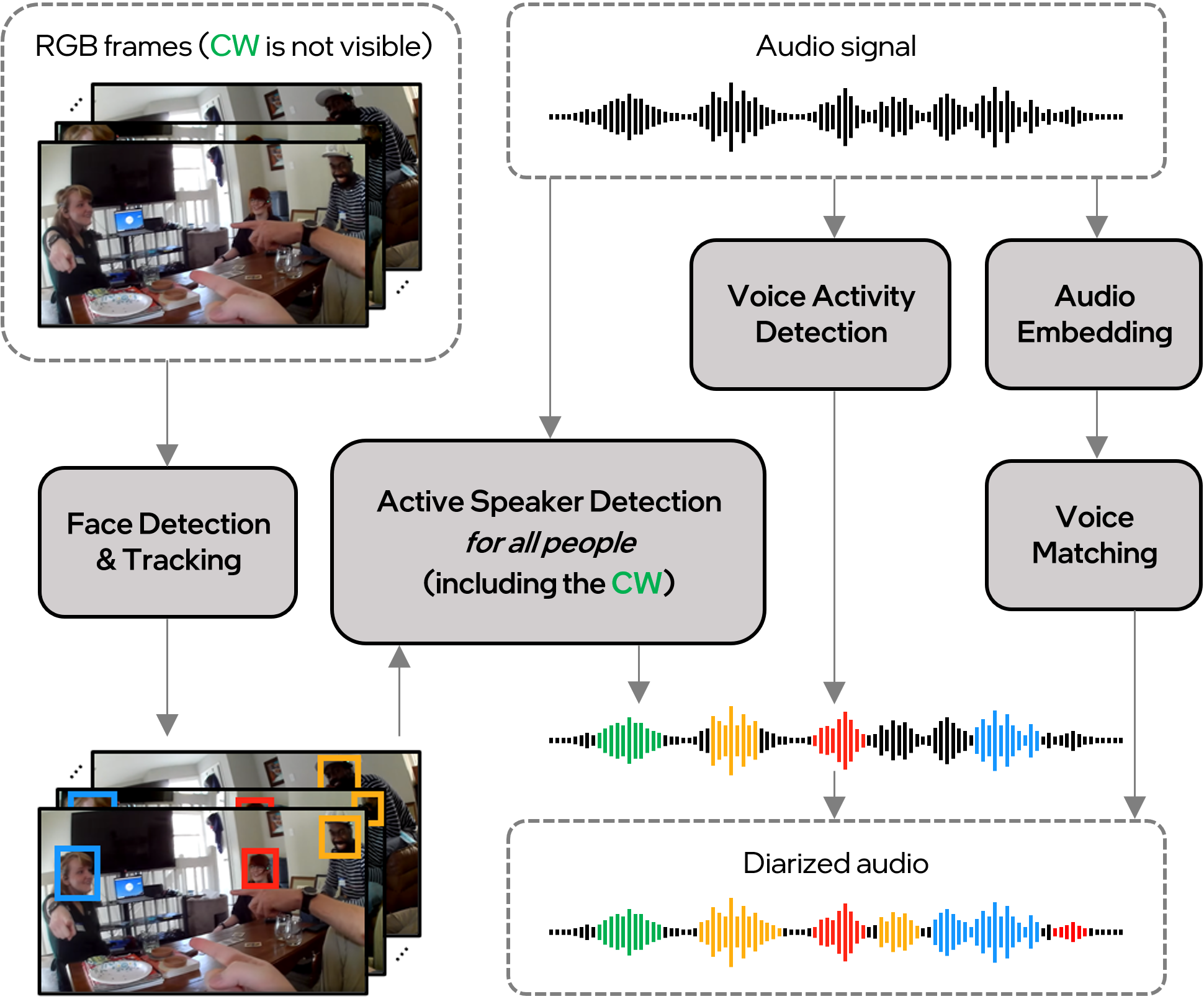}
  \caption{An illustration of the overall framework for AVD. First, face regions of visible people in the scene are detected and connected over time. Second, all active speakers, including the CW, are detected. Unlike previous approaches that rely on two separate models for the visible speakers and the (non-visible) CW, \ours{} can detect every speech activity using a single, unified framework. Finally, voice activity detection and voice matching modules refine the initial diarization by reducing false negatives and false positives, respectively. For these refinement steps, we use the same techniques presented in last year's submission~\cite{min2022intel}. We use four colors (red, blue, yellow, and green) to represent four different speaking identities including CW. Best viewed in color.}
  \label{fig:overview}
\end{figure}

\begin{figure}[t!]
\centering
\includegraphics[width=\linewidth]{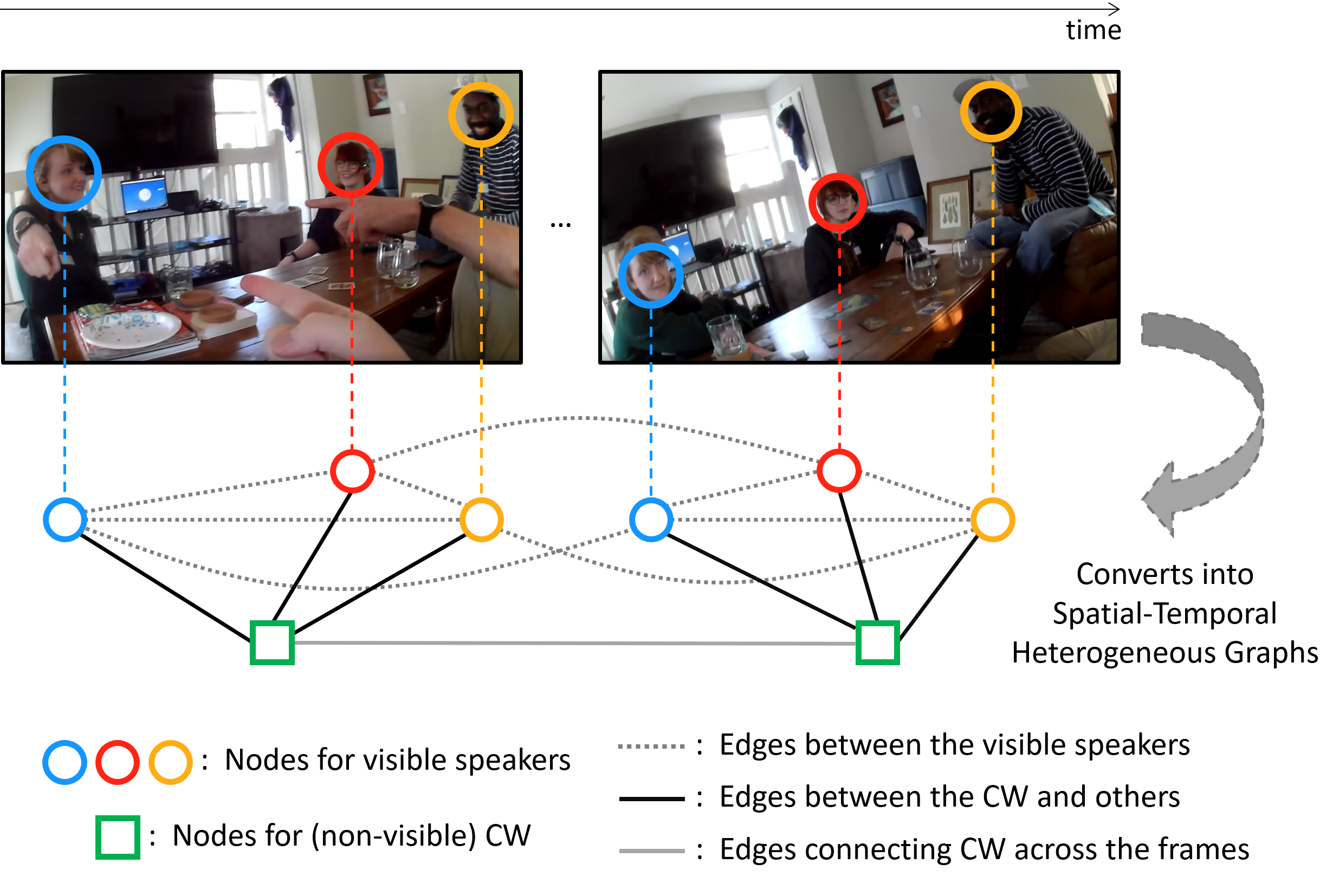}
  \caption{\ours{} converts a video into spatial-temporal heterogeneous graphs, where each node corresponds to a person and the edges represent spatial or temporal interactions between the speakers. Specifically, there are two types of nodes: the nodes for visible speakers are associated with audio-visual features, whereas the node features for the CW are audio-only. Accordingly, we can think of three types of edges depending on their connecting nodes. We can detect all the speakers including the CW, by solving binary node classification in the constructed heterogeneous graphs.}
  \label{fig:sthg}
\end{figure}

To this end, we propose a novel method based on Spatial-Temporal Heterogeneous Graph learning (\ours{}). As shown in Figure~\ref{fig:overview}, our key innovation is that we detect every speech activity using a single, unified framework. This unified method can jointly model all the speakers in the scene, including the CW, which significantly improves the speech detection performance compared to previous approaches. Figure~\ref{fig:sthg} illustrates how \ours{} constructs spatial-temporal heterogeneous graphs from egocentric videos. Since the CW's face is not visible in the scene, there are two types of nodes: the nodes for the visible speakers, and other nodes for the non-visible CW. The node features for visible speakers are audio-visual, whereas those of the CW are audio-only. Accordingly, we can think of the three types of spatial and temporal interactions between the nodes, which are described by three different types of edges in Figure~\ref{fig:sthg}. By using the constructed heterogeneous graphs, \ours{} jointly models and detects every speech activity in a unified way.

We show the effectiveness of \ours{} by performing experiments on the Ego4D dataset~\cite{grauman2022ego4d}. Specifically, we utilize the same set of face tracks, audio features for the CW, and audio-visual features for other speakers that were used in last year's submission~\cite{min2022intel}. Using the proposed spatial-temporal heterogeneous framework, \ours{} increases the mAP@0.5 score of active speaker detection (ASD) from 60.7\% to 63.7\% (and increases the mAP score of ASD from 71.3\% to 75.7\% when assuming a perfect face detector). More importantly, it also raises the detection mAP score of CW's voice activity from 80.4\% to 85.6\%. We want to note that these performance boosts are solely derived from the joint modeling ability of \ours{}. By fully leveraging the inter-speaker context between the CW and the others, the detection scores for both types of speakers are increased all together.

Our final method obtains 61.1\% DER on the test set, improved from 65.9\% that last year's winner acquired. Our submission achieved 1st position in the Ego4D Challenge 2023 leaderboard. Additionally, we address the Speech Transcription task of the challenge using the final output of \ours{}. We show that a competitive speech recognition performance can be achieved just by applying the off-the-shelf speech recognition system to the diarized speech segments.

\section{Method}
\label{sec:method}

The entire framework consists of six components as presented in Figure~\ref{fig:overview}. Among them, we mainly describe our innovations in ASD. We generally follow the approaches used by last year's winner~\cite{min2022intel} for other components.

\subsection{Heterogeneous Graph Construction} \label{subsec:gc}
Our heterogeneous graph construction process is illustrated in Figure~\ref{fig:sthg}. For each frame of a video, \ours{} connects all the nodes regardless of their identities to model their spatial context. In addition, the nodes corresponding to the same person are connected across the frames to model the context of temporal progression. For this temporal modeling, we use the same techniques used in previous work~\cite{min2022learning}. The two types of connections, which are actually represented by three different types of edges in Figure~\ref{fig:sthg}, are expressive enough to jointly model the spatial-temporal interactions between the CW and others. Furthermore, the constructed graphs are very sparse (graph density $<$ 3\%), which makes it possible to train and evaluate \ours{} within small memory and computation budget. We used a single GPU (TITAN V) for all the experiments.

\subsection{Active Speaker Detection} \label{subsec:asd}
We convert the problem of ASD for all people including the CW into a binary node classification task on the constructed heterogeneous graphs. During the training process, a graph neural network (GNN) is optimized on the ground-truth binary labels indicating whether a certain node is speaking or not. It consists of three layers with two different methods of neighborhood aggregation, following the previous work~\cite{min2022learning}.

\subsection{Voice Activity Detection} \label{subsec:vad}
Following the previous year's winner~\cite{min2022intel}, we use an off-the-shelf voice activity detection (VAD) model called Silero~\cite{Silero_VAD} (we refer to it as SilVAD). We confirm that SilVAD reduces the false positives from the speech detections of the CW when it is used for post-processing.

\begin{table*}[t]
\centering
\resizebox{\linewidth}{!}{
\begin{tabular}{L{2.6cm}|C{3.4cm}|C{3.4cm}|C{3.4cm}}
\toprule
\, \textbf{ASD Model} & \textbf{ASD mAP(\%)}$\uparrow$ & \textbf{ASD mAP@0.5(\%)}$\uparrow$ & \textbf{AVD DER(\%)}$\downarrow$ \\ \midrule
\, RegionCls~\cite{grauman2022ego4d} & - & 24.6 & 80.0 \\
\, TalkNet~\cite{tao2021someone} & - & 50.6 & 79.3 \\
\, SPELL~\cite{min2022learning,min2022intel} & 71.3 & 60.7 & 66.6 \\ \midrule
\, \ours{} (Ours) & \textbf{75.7} & \textbf{63.7} & \textbf{59.4} \\
\bottomrule
\end{tabular}
}
\caption{ASD and AVD performance comparisons on the validation set of the Ego4D dataset. \textbf{We report two metrics to evaluate ASD performance: mAP quantifies the ASD results by assuming that the face bound-box detections are the ground truth (i.e. assuming the perfect face detector), whereas mAP@0.5 quantifies the ASD results on the detected face bounding boxes (i.e. a face detection is considered positive only if the IoU between a detected face bounding box and the ground-truth exceeds 0.5).} We compute mAP@0.5 by using the official evaluation tool provided by Ego4D\protect\footnotemark. For AVD, we report DER (diarization error rate): a lower DER value indicates a better AVD performance.}
\label{tab:asdavd}
\end{table*}

\subsection{Speech Recognition} \label{subsec:sr}
We additionally address the Speech Transcription task of the Ego4D Challenge~\cite{grauman2022ego4d}. After having the final output of \ours{} in the form of the diarized audios, we apply the off-the-shelf speech recognition method to each speech segment. In particular, we use one of the ESPNet2 models~\cite{watanabe2018espnet} that is pre-trained on the CHiME-5 dataset~\cite{watanabe2020chime}. This dataset consists of real-world recordings that were captured in everyday home environments, which we find to be similar to the Ego4D scenarios.

\footnotetext{\url{https://github.com/EGO4D/audio-visual/tree/main/active-speaker-detection/active_speaker/active_speaker_evaluation}}

\begin{table}[t]
\centering
\resizebox{\linewidth}{!}{
\begin{tabular}{L{5.5cm}|C{1.75cm}}
\toprule
\textbf{Method} & \textbf{mAP(\%)}$\uparrow$ \\ \midrule
Energy filtering + Audio Matching~\cite{grauman2022ego4d} & 44.0 \\
Spectrogram + ResNet-18~\cite{he2016deep,grauman2022ego4d} & 72.0 \\
SPELL~\cite{min2022learning,min2022intel} & 80.4 \\ \midrule
\ours{} (Ours) & \textbf{85.6} \\
\bottomrule
\end{tabular}
}
\caption{Performance comparison of our method with the baselines used in the Ego4D paper~\cite{grauman2022ego4d} on the validation set of the Ego4D dataset. We report the mAP (mean average precision) of the detected CW's speaking activities.}
\label{tab:pcwvad}
\end{table}

\section{Experiments}
\label{sec:exp}

\subsection{Comparison with the state-of-the-art} \label{subsec:sota}

We summarize the ASD and AVD performance comparisons of \ours{} with other state-of-the-art approaches on the validation set of the Ego4D dataset~\cite{grauman2022ego4d} in Table~\ref{tab:asdavd}. Our method brings significant performance gains for both ASD and AVD, which demonstrate the effectiveness of the proposed spatial-temporal heterogeneous graph learning. Moreover, as shown in Table~\ref{tab:pcwvad}, we observed that \ours{} raises the mAP score of CW's speaking activity detection from 80.4\% to 85.6\% even when using the same set of features as last year's submission. This is because \ours{} can fully leverage the inter-speaker context between the CW and the other visible speakers.

\subsection{Benefit of VAD} \label{subsec:asdavd}

In Table~\ref{tab:pp}, we show the performance of \ours{} under different post-processing conditions on the validation set of the Ego4D dataset~\cite{grauman2022ego4d}. We can observe that post-processing improves the AVD performance when it is applied to CW's detection only. This observation agrees well with the previous work~\cite{min2022intel}.

\begin{table}[t!]
\centering
\begin{tabular}{C{1.9cm}C{1.9cm}|C{1.9cm}}
\toprule
\multicolumn{2}{c|}{\textbf{Post-processing w/ SilVAD}} & \multirow{2}{*}{\textbf{DER(\%)}$\downarrow$} \\[2pt] CW & Others & \\ \midrule
& & 60.2 \\
$\checkmark$ & & \textbf{59.4} \\
& $\checkmark$ & 63.1 \\
$\checkmark$ & $\checkmark$ & 62.5 \\
\bottomrule
\end{tabular}
\caption{Performance of \ours{} under different post-processing conditions on the validation set of the Ego4D dataset. We confirm that post-processing boosts the performance when it is applied to the CW's detections only.}
\label{tab:pp}
\end{table}

\subsection{Final AVD performance} \label{subsec:final}

\begin{table}[t]
\centering
\begin{tabular}{L{4.2cm}|C{2cm}}
\toprule
\textbf{Method} & \textbf{DER(\%)}$\downarrow$ \\ \midrule
Baseline using TalkNet~\cite{tao2021someone} & 73.3 \\
SPELL~\cite{min2022learning,min2022intel} & 65.9 \\ \midrule
\ours{} (Ours) & \textbf{61.1} \\
\bottomrule
\end{tabular}
\caption{Final AVD performance on the test set of the Ego4D dataset. Our method significantly outperforms the previous submissions.}
\label{tab:final}
\end{table}

We report the final AVD performance of our method in Table~\ref{tab:final}. When compared to the leading state-of-the-art method~\cite{min2022learning}, \ours{} achieves 4.8\% lower DER on the test set, attaining 1st place in Ego4D Challenge 2023.

\subsection{Speech Transcription} \label{subsec:speech}
Additionally, we apply the off-the-shelf speech recognition model on the diarized segments and obtain 78.3\% WER (word error rate) on the validation set. For the Speech Transcription task of the challenge, we achieved 78.6\% WER, which outperforms the baseline provided by the challenge organizers (112.1\% WER). We believe that fine-tuning the speech recognition model will lead to a significant reduction in errors.

We also observe that the ground-truth time segments for the speech transcription are not temporally aligned with the ones for diarization. This is probably because the transcript annotating process was done separately as described in the Ego4D paper~\cite{grauman2022ego4d}. Therefore, we believe that fine-tuning the ASD and VAD models will also reduce the errors by finding speech segments that are better aligned with the transcript annotations.

\section{Conclusion}
\label{sec:concl}

We have presented \ours{} as an advanced framework for audio-visual diarization. The main idea is simple and effective: it models all the speakers, whether they are visible or not, in a unified structure of heterogeneous graphs. Consequently, we can jointly detect all the speakers including the CW, and fully leverage the inter-speaker context between the CW and others. \ours{} significantly outperforms previous AVD methods and provides a competitive speech transcription outcome.

{\small
\bibliographystyle{ieee_fullname}
\bibliography{main}
}

\end{document}